\documentclass{article}
\usepackage{ijcai24}

\usepackage{times}
\usepackage{soul}
\usepackage{url}
\usepackage[hidelinks]{hyperref}
\usepackage[utf8]{inputenc}
\usepackage[small]{caption}
\usepackage{graphicx}
\usepackage{amsmath}
\usepackage{amsthm}
\usepackage{booktabs}
\usepackage{algorithm}
\usepackage{algorithmic}
\usepackage[switch]{lineno}

\usepackage{enumitem}
\usepackage{color}

\def\tony#1{\textcolor{black}{#1}} 
\def\fj#1{\textcolor{black}{#1}} 

\title{
\title{Reframing Spatial Reasoning Evaluation in Language Models:\\ A Real-World Simulation Benchmark for Qualitative Reasoning}

}

\author{
Fangjun Li$^1$
\and
David C. Hogg$^1$\and
Anthony G. Cohn$^{1,2}$
\affiliations
$^1$School of Computing, University of Leeds, UK \\
$^2$Alan Turing Institute, UK
\emails
\{scfli, d.c.hogg, a.g.cohn\}@leeds.ac.uk
}

\begin{document}

\maketitle

\begin{abstract}

Spatial reasoning plays a vital role in both human cognition and machine intelligence, prompting new research into language models' (LMs) capabilities in this regard. However, existing benchmarks reveal shortcomings in evaluating qualitative spatial reasoning (QSR). These benchmarks typically present oversimplified scenarios or unclear natural language descriptions, hindering effective evaluation. 
We present a novel benchmark for assessing QSR in LMs, which is grounded in realistic 3D simulation data, offering a series of diverse room layouts with various objects and their spatial relationships. This approach provides a more detailed and context-rich narrative for spatial reasoning evaluation, diverging from traditional, toy-task-oriented scenarios.
Our benchmark encompasses a broad spectrum of qualitative spatial relationships, including topological, directional, and distance relations. These are presented with different viewing points, varied granularities, and density of relation constraints to mimic real-world complexities. A key contribution is our logic-based consistency-checking tool, which enables the assessment of multiple plausible solutions, aligning with real-world scenarios where spatial relationships are often open to interpretation.
Our benchmark evaluation of advanced LMs reveals their strengths and limitations in spatial reasoning. They face difficulties with multi-hop spatial reasoning and interpreting a mix of different view descriptions, pointing to areas for future improvement.

\end{abstract}

\section{Introduction}

In recent years, advancements in language models \cite{OpenAI2023GPT4TR} \cite{touvron2023llama} have significantly improved their capabilities in understanding and reasoning with textual information \cite{li2022ontology}. 
 However, promoting these models' ability to process and reason about spatial relationships remains a complex challenge \cite{bang2023multitask} \cite{cohn2023dialectical}.
Spatial reasoning, a critical component of human cognition, involves understanding and navigating the relationships between objects in space \cite{cohn2008qualitative} 
\cite{alomari2022online}. 
Existing benchmarks like bAbI \cite{weston2016towards}, StepGame \cite{shi2022stepgame}, SpartQA\cite{mirzaee2021spartqa}, and SpaRTUN \cite{mirzaee-kordjamshidi-2022-transfer} have significantly contributed to the field, yet they exhibit limitations in representing the complexity and naturalness found in real-world spatial reasoning.

In this paper, we conduct an extensive analysis of task complexity and limitations in four widely used datasets for textual spatial reasoning evaluation. 
bAbI and StepGame, originating from simplified, toy-like tasks, utilize grid-based environments with fixed distances and angles for spatial relations. This approach for constructing spatial reasoning data, while ensuring unique solutions, oversimplifies the tasks, failing to capture the complexity of spatial relationships in the real world. Moreover, the primary challenge in StepGame lies in constructing a chain of objects from multiple shuffled relations, overshadowing the spatial reasoning aspect. Our previous research indicates that GPT-4 excels in the spatial reasoning aspects of relation mapping and coordinate calculation needed for this task once the chain is established.

On the other hand, SpartQA and SpaRTUN, which cover a wider range of spatial relationships, 
do not always contain clear and fluent language descriptions. Common issues observed include complex object descriptions and disordered relational sequencing. Objects are described using a combination of color, size, and shape. This level of detail complicates the narrative, shifting the focus away from spatial reasoning and towards deciphering the object descriptions. The disordered relational sequencing hinders the understanding of the core spatial problem, adding unnecessary complexity.


In response to the limitations of current benchmarks in qualitative spatial reasoning, this paper introduces a new, more comprehensive benchmark to evaluate LMs' abilities in this domain.
Our benchmark seeks to present more naturally described stories, employing language that is easily understandable and processable by both humans and LMs. We aim to move away from overly logical expressions and toward narratives that mirror everyday communication. To achieve this goal, the scenarios for our benchmark are sourced from 3D simulation data rather than 
toy 
tasks, encompassing a variety of room layouts with diverse objects, each annotated with specific attributes. This approach allows each scenario to showcase a distinct arrangement of everyday objects.
During data creation, the placement of objects, their layout, and their spatial relationships with other objects are determined. This information forms the basis for generating stories, questions, and answers for each instance.


\begin{figure}[t]
    \centering
    \includegraphics[width=0.48\textwidth]{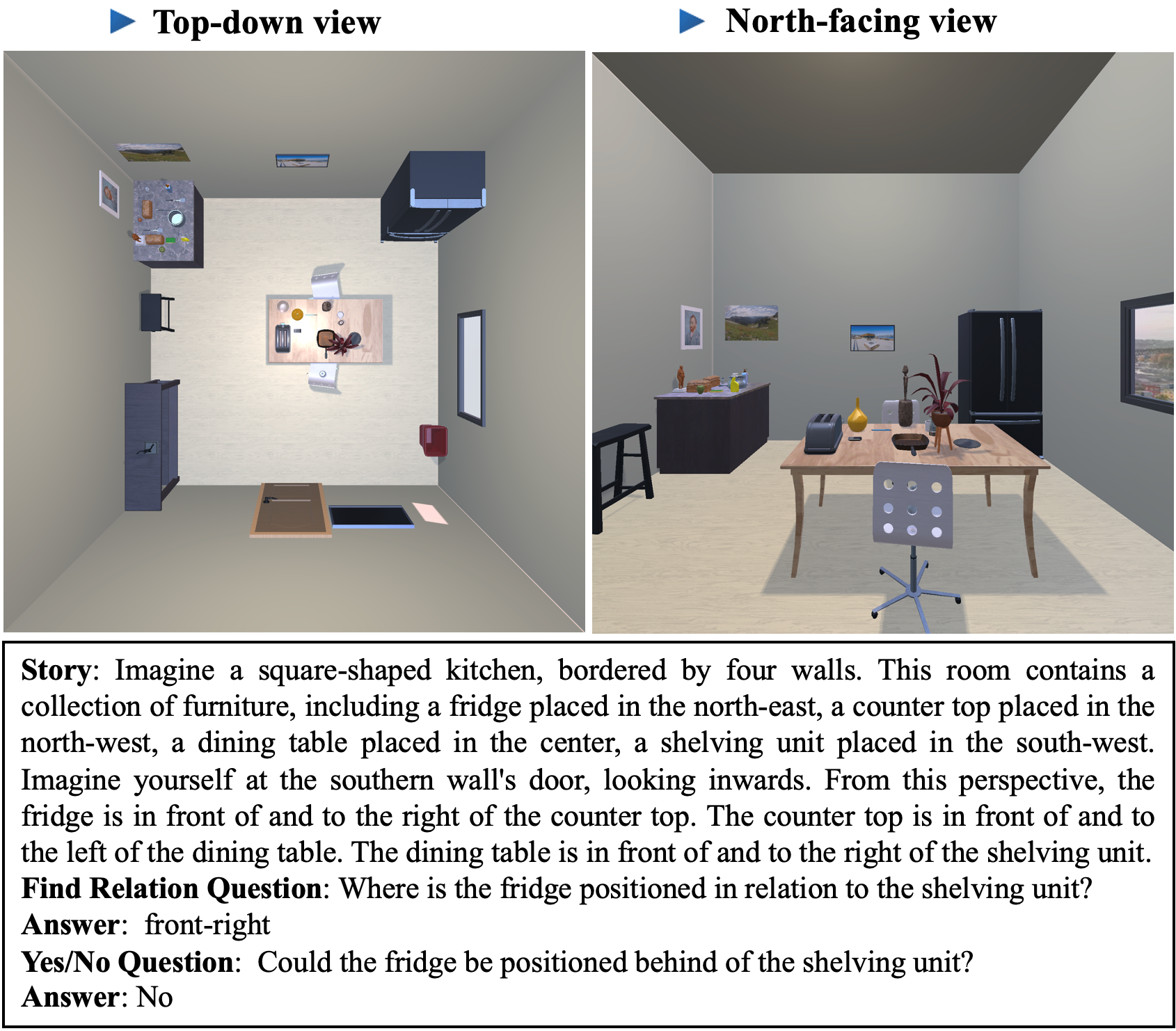}
    \caption{One test instance in our benchmark, consisting only of text for evaluating LMs. The accompanying images are for visualization
    \fj{but could be used to test multi-modal LLMs.}
    }
    \label{fig:example}
\end{figure}

Recognizing that spatial reasoning often yields multiple plausible solutions, we focus on assessing the consistency of LMs' answers within the given constraints rather than seeking a single `correct' answer. This approach aligns with the real-world nature of spatial reasoning, where multiple interpretations are often valid.

Finally, we evaluate some LLMs' performance on our benchmark, to offer a more nuanced and comprehensive evaluation of LLMs' qualitative spatial reasoning ability.
According to our results, GPT-4 shows superior capability in spatial reasoning tasks across various settings. 
All models face challenges in reasoning about spatial relations between objects as multi-hop spatial reasoning complexity increases. However, there is a clear trend toward improved performance as the story's constraint graph becomes more complete.

This paper presents several contributions to the field of QSR evaluation, particularly in the context of LM performance. These contributions are as follows:
\begin{itemize}[leftmargin=*]
    \item Comprehensive analysis of existing benchmarks. We provide an in-depth analysis of the complexity and limitations inherent in current spatial reasoning benchmarks. 
    \item Constructing a more natural and realistic benchmark by developing scenarios derived from 3D simulation data, offering a diverse series of data, each varying in the granularity of relationships and the selection of relational constraints.
    \item Introduction of a logic-based consistency checking tool for evaluation, which evaluates whether spatial relations predicted by LMs are feasible, given the set constraints.
    \item Detailed evaluation of LLMs' spatial reasoning abilities. By applying our benchmark to test various LMs, we provide a refined assessment of their capabilities in QSR. 
\end{itemize}

Overall, these contributions \fj{advance LM evaluation for spatial reasoning}, aligning more closely with real-world scenarios and human cognitive processes.

\section{Analysis of Existing QSR in Text Datasets/Benchmarks}
Representative benchmarks like bAbI, StepGame, SpartQA, and SpaRTUN focus on spatial reasoning. They involve tasks where models are required to infer new spatial relation\fj{s} from provided facts or check the consistency of relations.

\subsection{bAbI}

The bAbI benchmark \cite{weston2016towards}, featuring a collection of synthetic tasks, was crafted to evaluate learning algorithms in terms of their text understanding and reasoning abilities. 
Among its 20 tasks, Tasks 17 and 19 are specifically designed for spatial reasoning evaluation.

\begin{figure}[htb]
    \centering
    \includegraphics[width=0.48\textwidth]{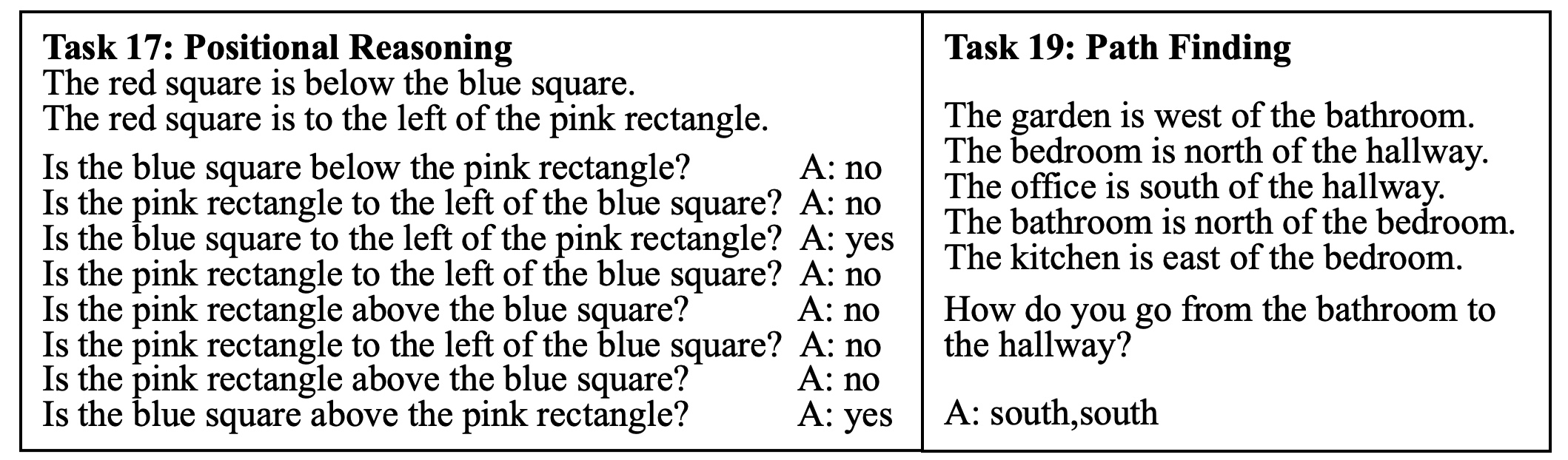}
    \caption{Examples of Task 17 and Task 19 from the bAbI's envalid-10k dataset version.}
    \label{fig:babi}
\end{figure}

Task 17 tests LM\fj{s'} ability to understand and reason about relative spatial relations `left', `right', `above', and `below'. 
The task operates within a 5x5 grid environment. In this structured setting, three entities are sequentially positioned at specific nodes. The placement of each entity is determined by its spatial relation to the adjacent nodes.
The narratives distinguish three entities based on their color and shape. Each example can include up to 10 sentences - 2 describing spatial relations between two pairs of objects and 8 for generating questions about a different pair, as illustrated in Figure \ref{fig:babi}. These questions are structured in a yes/no format, with answers based on the entities' actual positions on the grid.

Task 19 is centered around identifying paths between specified objects, utilizing the four cardinal directions: north, south, east, and west. These objects are described as various locations, such as bedrooms and bathrooms. In the `en-valid-10k' version of bAbI\footnote{https://www.kaggle.com/datasets/roblexnana/the-babi-tasks-for-nlp-qa-system}, each story typically includes 5 sentences related to spatial relations: 2 effectively describing the path and 3 serving as decoys, as shown in Figure \ref{fig:babi}. The task's challenge lies in mapping out a sequential path from the start entity to the end entity. The inclusion of decoy sentences adds a layer of complexity to the task.

The bAbI tasks, designed as simplified `toy tasks', have limitations in \fj{testing} spatial reasoning. They restrict spatial relations to basic cardinal directions north, south, west, and east (also referred to as above, below, left, and right in task 17) with set distances and angles, lacking the complexity and ambiguity of real-world spatial scenarios. Additionally, using a single template for each relation may not adequately challenge a model's understanding and reasoning in more nuanced, context-rich environments. Consequently, while useful for basic training, bAbI tasks may not fully 
test or
equip models for the intricacies of real-world spatial reasoning.


\subsection{StepGame}

\begin{figure}[tb]
    \centering
    \includegraphics[width=0.48\textwidth]{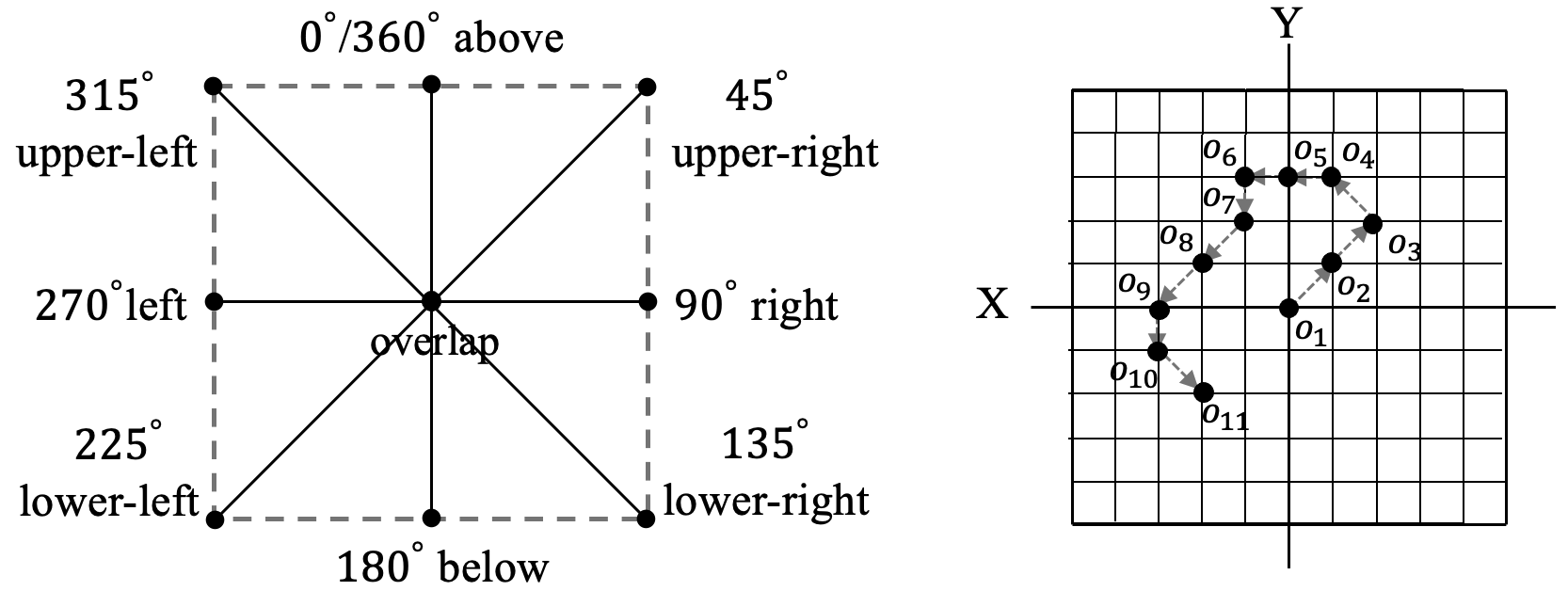}
    \caption{Illustration of directional spatial relationships and test instance constraint chain building process in StepGame.}
    \label{fig:stepgame}
\end{figure}

Building upon bAbI, the StepGame benchmark \cite{shi2022stepgame} utilizes a grid-based system and introduces higher complexity in three key aspects:

\begin{itemize}[leftmargin=*]
    \item An expanded set of directional spatial relations is included, encompassing eight relations: top (north), down (south), left (west), right (east), top-left (north-west), top-right (north-east), down-left (south-west), and down-right (south-east). Each is defined by a unique angle and distance.
    These relations can be visually illustrated on a grid, as shown in the left diagram of Figure \ref{fig:stepgame}, with the inclusion of an `overlap' relation for overlapping object locations.
    \item Enhanced multi-hop reasoning challenges: Moving beyond the 4-hop reasoning in bAbI, StepGame increases the complexity to span 1-hop to 10-hop sequences. The right diagram of Figure \ref{fig:stepgame} illustrates the sequential building of relational constraints, based on $k$, the number of relationships. This produces a chain of constraints linking objects in a direct path from $o_1$ to $o_2$, continuing through to $o_{n+1}$.
    \item Employing richer, crowdsourced narratives describing eight possible spatial relations between two entities, which serve as the basis for generating \fj{story-question pairs}. 
\end{itemize}

The spatial configuration used in StepGame introduces limitations that may affect the evaluation of LMs' spatial reasoning abilities.
Commonsense human understanding does not confine directional relationships to strict distance or angular constraints.
For example, when we say `A is east of B' in a two-dimensional framework, it simply means that the x-coordinate of A, denoted as $x_A$, is larger than that of B, $x_B$. This does not necessarily dictate that $x_A$ should exceed $x_B$ by an exact value or align with a specific angle, such as a 1-unit difference or a $90^\circ$ angle.

StepGame's design yields unique solutions for all instances, but with limited complexity (as depicted in the Appendix).
Prior research \cite{li2024advancing}
indicates that the most challenging aspect for LLMs in this task is constructing the object-linking chain from shuffled relations, rather than the spatial reasoning component itself. When provided with a pre-constructed reasoning chain, GPT-4 demonstrates remarkable proficiency in handling such reasoning tasks. 


\subsection{SpartQA, SpaRTUN:}
SpartQA \cite{mirzaee2021spartqa} and SpaRTUN \cite{mirzaee-kordjamshidi-2022-transfer} start from 2D images featuring objects (rectangle, triangle, square) distributed across distinct square blocks (scenes). They extend beyond mere directional spatial relationships to include Region Connection Calculus 8 (RCC-8) \cite{randell1992spatial} and distance (near and far). SpaRTUN is an updated version of SpartQA-Auto and contains more relation types and rules.

\fj{Unlike} the previous two grid-based benchmarks, SpartQA and SpaRTUN's define spatial relations using a square boundary framework. Each spatial relation is determined by the $(x,y)$ coordinates of the lower-left points of the square boundary boxes of two objects and the size of these boxes.

\begin{itemize}[leftmargin=*]
    \item For object-to-object relations, EC, NEAR, FAR, LEFT / RIGHT, ABOVE / BELOW are considered;
    \item For object-to-scene relations, TPP / TPPi, and NTPP / NTPPi are considered;
    \item For scene-to-scene relations, DC, EC, PO, TPP / TPPi, and NTPP / NTPPi are considered.
\end{itemize}


The scene description was generated from the selected story triplets using context-free grammar (CFG). They increase the variety of spatial expressions by using a vocabulary of various entity properties and relation expressions. They map the relation types and the entity properties to the lexical forms from a specifically collected vocabulary.

\begin{figure}[tb]
    \centering
    \includegraphics[width=0.48\textwidth]{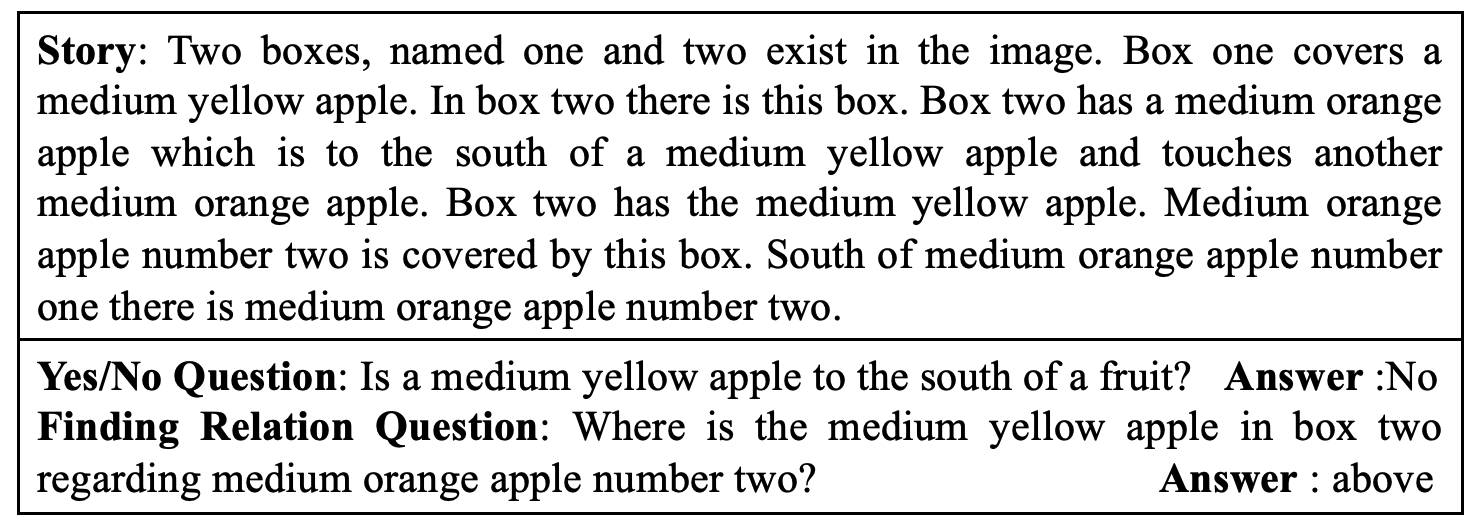}
    \caption{A test example in SpaRTUN.}
    \label{fig:spartun}
\end{figure}

Although these two benchmarks include rich spatial relationships, they struggle to provide effective descriptions. They use simple syntax and word choice but \fj{lack logical flow and content clarity,} particularly in two aspects:
\begin{itemize}[leftmargin=*]
    \item The spatial relations are described as a sequence of randomly selected story triplets, which deviates from the typical human approach to describing a scene. In the example from Figure \ref{fig:spartun}, a more natural human description would typically start with outlining the relationships between two boxes, followed by detailing the contents of each box, and then explaining the relations between the objects. However, in their narrative structure, there is a lack of an initial summary of the objects contained in each box, with objects being introduced individually and somewhat disjointedly. Additionally, the narrative places the object-to-box relationships prior to the box-to-box relationships, which further diverges from the typical human method of spatial description, leading to potential confusion in understanding the overall spatial layout.
    
    \item The excessive use of detailed and repetitive entity naming, involving terms like `medium yellow apple', `medium orange apple number one', and `medium orange apple number two', results in overly lengthy text. This verbosity transforms a simple description such as `South of A is B' into a more convoluted one like `South of medium orange apple number one is medium orange apple number two'. Such complexity not only adds confusion but also shifts the focus from understanding the spatial relationship to deciphering which specific object is being referred to. This can make it hard for readers to grasp the intended spatial relationships and hinder smooth comprehension.
\end{itemize}

Consequently, the narrative's lack of smooth flow in textual descriptions makes it difficult for both LMs and humans to form a clear mental image of the entire scene and to grasp information about specific objects in question. This complexity hinders the LMs from engaging in spatial reasoning effectively and drawing conclusive answers based on the limited information presented.






\section{Data Generation Framework}

\subsection{Problem Definition}

We focus on constraint satisfaction problems (CSP), defined by a set of variables $V$ defined over a domain $D$  and a collection of constraints $\theta$. The goal is to find a specific instantiation where all constraints in $\theta$ are simultaneously satisfied.
We particularly emphasize binary constraints, which simultaneously restrict the domain of two variables. An example of this is  `The desk is placed in front of the sofa.'

\fj{One instance of spatial reasoning problem} can be conceptualized as a constraint network framework:
\fj{c}onsider a network comprising \( n \) spatial variables \(V = \{o_1, ..., o_n\}\) within a domain \( D^n \). In this network, each node is identified by a variable \( o_i \) or by the variable's index \( i \), and each directed edge is marked with a binary relation constraint. We use the notation  \( r{ij}\) to denote the relation that constrains the pair of variables \( \langle o_i, o_j \rangle \).
One relation constraint in \( \theta \) can thus be denoted as \( r_{ij}(o_{i}, o_{j}) \) or \((o_i, r_{ij}, o_j) \).

Given a set of \( k \) relations and a query \( (o_{a},r_{ab},o_{b}) \), LMs are tasked with predicting the relation \( r_{ab} \). If all constraints present in the story, including the predicted relation constraint  \( (o_{a},r_{ab},o_{b}) \), can be simultaneously satisfied, we consider the prediction to be an effective solution.


\subsection{Data Generation Process}

Our benchmark data encompasses a range of configurations, each aligning with specific elements of the constraint network. These configurations are denoted by the tuple \( \langle n, d, m, p \rangle \), where:
\begin{itemize}[leftmargin=*]
    \item \( n \) is the number of objects used to form the story in the scene, as is established through the process in Section \ref{decide_p}.
    \item \( d \) 
    \fj{is the number of square tiles in a } 
    \( width \times length \) tessellation whose centres define possible positions for the centres of objects on the floor plane. \tony{In the dataset, $width$ and $lenghth$ are always equal, yielding square rooms.}
    
    \item \( m \) is the number of binary constraints over $n$ objects, set by the method described in Section \ref{decide_p}. 
    The maximum possible number of constraints on \( n \) variables is  \( \frac{n(n-1)}{2} \), under which each variable is constrained \fj{by} all other variables and the graph is a complete graph, \fj{i.e., an n-clique}.
    \item \( p \) is the constraint tightness. For unary constraints, \( p \) ranges from 0 to \( d \), and for binary constraints, from 0 to \( d \times d \). Here, \( d \) is the domain size for one variable, \( d \times d \) corresponds to the total possible pairs of values between two variables. For each binary constraint, the number of disallowed value pairs is calculated as \( p \times (d \times d) \).
    \( p \) is related to the types of constraints, as outlined in Section \ref{decide_m}. We analyse the constraint tightness in the Appendix.
\end{itemize}

All constructed constraint networks are transformed into \fj{a} textual format using the method outlined in Section \ref{decide_t}, specifically for the purpose of evaluating LMs.
Our test sets are available in varying sizes: \textbf{RoomSpace-100} includes a sample of 100 rooms.
\textbf{RoomSpace-1K} consists of 1,000 rooms, and \textbf{RoomSpace-10K} comprises 10,000 rooms.
The initial 100 rooms in RoomSpace-1K (ID 0-99) are identical to those in RoomSpace-100. Similarly, the first 1,000 rooms in RoomSpace-10K (ID 0-999) match those in RoomSpace-1K.


\subsection{Define House Scenes and Objects}
\label{decide_n}

\begin{figure}[tb]
    \centering
    \includegraphics[width=0.48\textwidth]{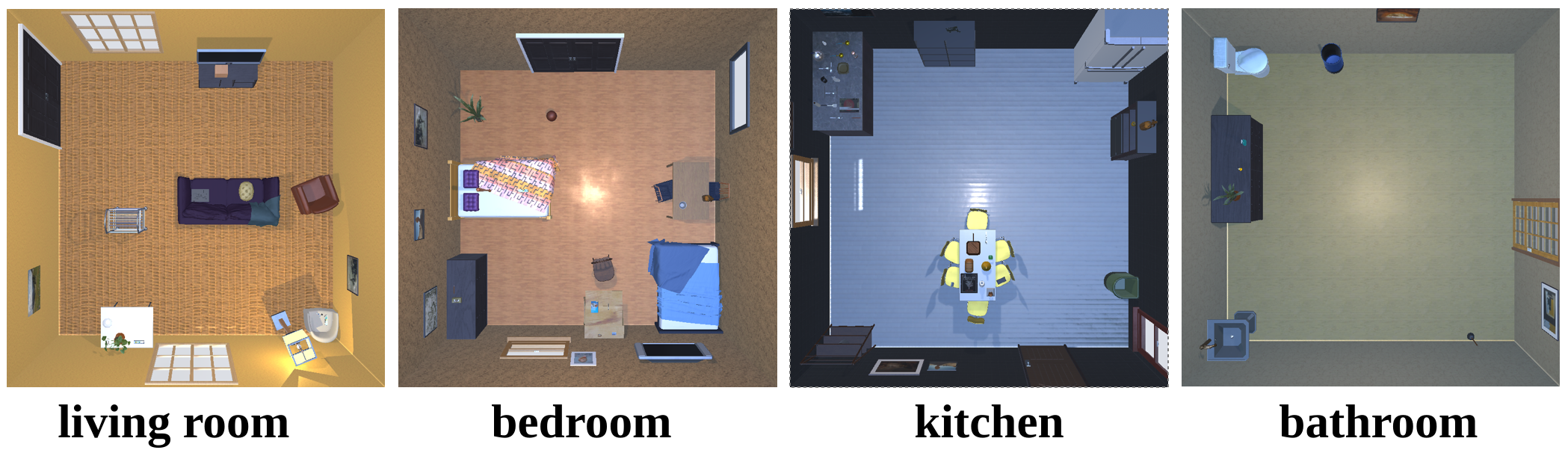}
    \caption{Sample scenes from our dataset showcasing four types of rooms in a top-down view.}
    \label{fig:images}
\end{figure}

We utilize the ProcTHOR \cite{deitke2022️} framework to create physics-enabled environments, which allow for the generation of a variety of virtual house environments. 
The initial ProcTHOR dataset includes simulated houses with multiple rooms. 
For our indoor setup, we adapt this to generate scenes within a single-room configuration to simplify the spatial reasoning challenges (see Figure \ref{fig:images} for examples).

Each room is uniformly square-shaped, enclosed by four walls (north, south, east, and west) that incorporate elements such as doors and windows. Despite this structural consistency, each room type is distinguished by
diverse configurations of household objects.






\subsection{Specify Spatial Relationships}
\label{decide_m}
We incorporate three types of spatial relations: topological, directional, and distance relations. These are utilized to detail the positioning of objects within rooms ($C_{l}$) and to define the relationships between objects ($C_{o}$).
The layout constraints, \(C_l\), are expressed as \((o_i, r_{i}, Room), i \in [1, n]\), and the inter-object constraints, \(C_o\), are formulated as \((o_i, r_{ij}, o_j), i \neq j\).

\subsubsection{Object Layout within Room}
We incorporate directional and topological spatial relationships to detail how objects are positioned within rooms. 

\paragraph{Directional Relations.}
The representation of  directional relations \tony{between objects} extended in 2D space,  we just use their central points.
As depicted in the left part of Figure \ref{fig:direction}, we divide the room into nine regions: North (N), West (W), East (E), South (S), Center (C), North-West (NW), North-East (NE), South-West (SW), and South-East (SE). The location of an object in a room is determined by the region in which the centre of its bounding box is situated.

\begin{figure}
    \centering
    \includegraphics[width=0.48\textwidth]{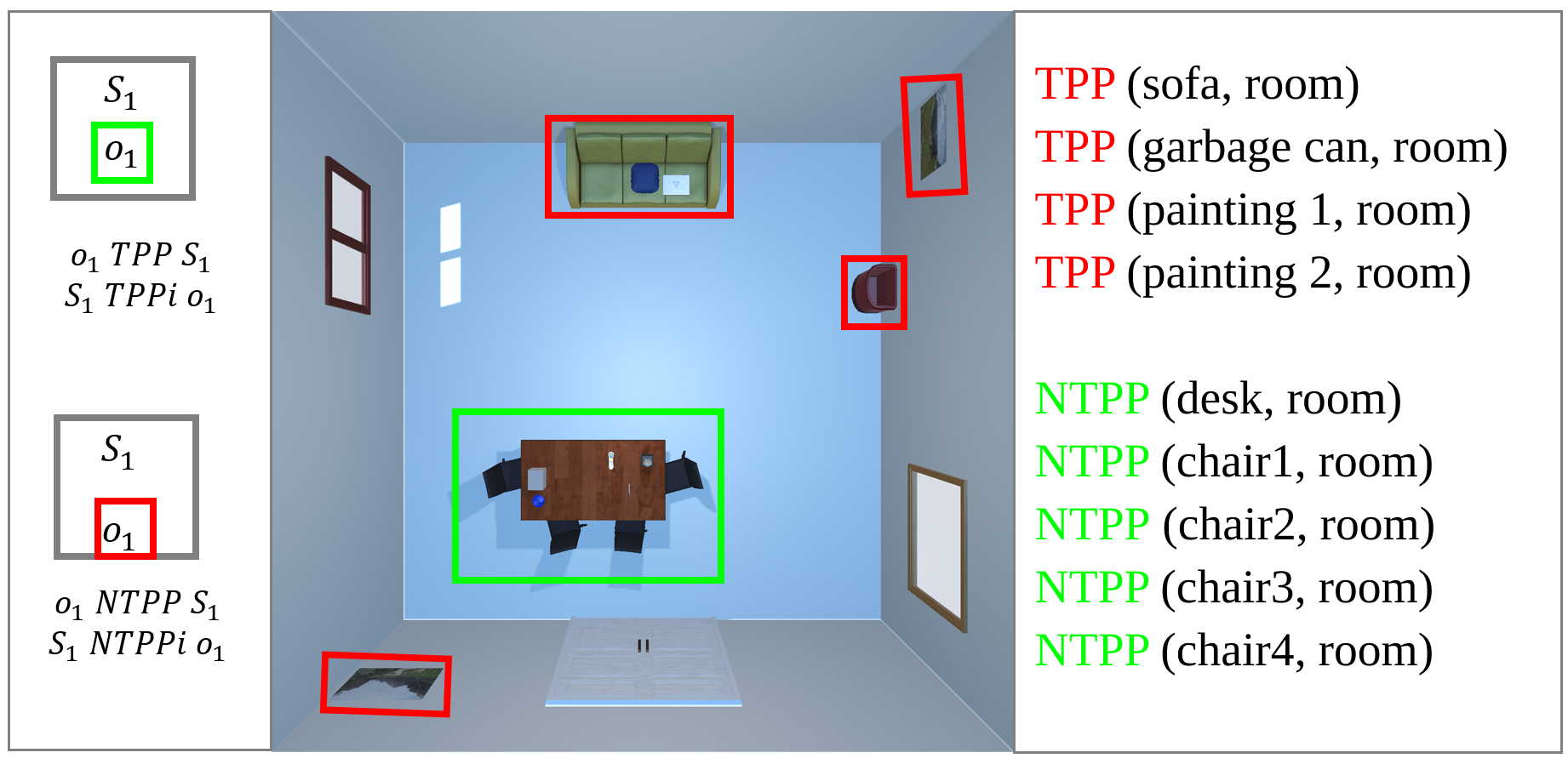}
    \caption{Illustration showcasing two topological spatial relations: TPP (in red, denoting objects touching the room's walls) and NTPP (in green, representing objects positioned inside the room's boundaries without touching the walls).
    }
    \label{fig:tpp}
\end{figure}

\paragraph{Topological Relations.} Two settings are considered:
\begin{itemize}[leftmargin=*]
    \item Uniform Inclusion. All objects are considered within the room, with no specific topological distinctions made.
    \item Tangential Proper Part (\textit{TPP}) and Non-Tangential Proper Part (\textit{NTPP}). Just record objects' topological relations to the wall, not the floor, as depicted in Figure \ref{fig:tpp}.
\end{itemize}


\subsubsection{Relations between Objects}
We define the relationships between any two objects using directional and distance-based spatial relations, determined by comparing the \( x \) and \( y \) coordinates of their centre points.

\begin{figure}
    \centering
    \includegraphics[width=0.48\textwidth]{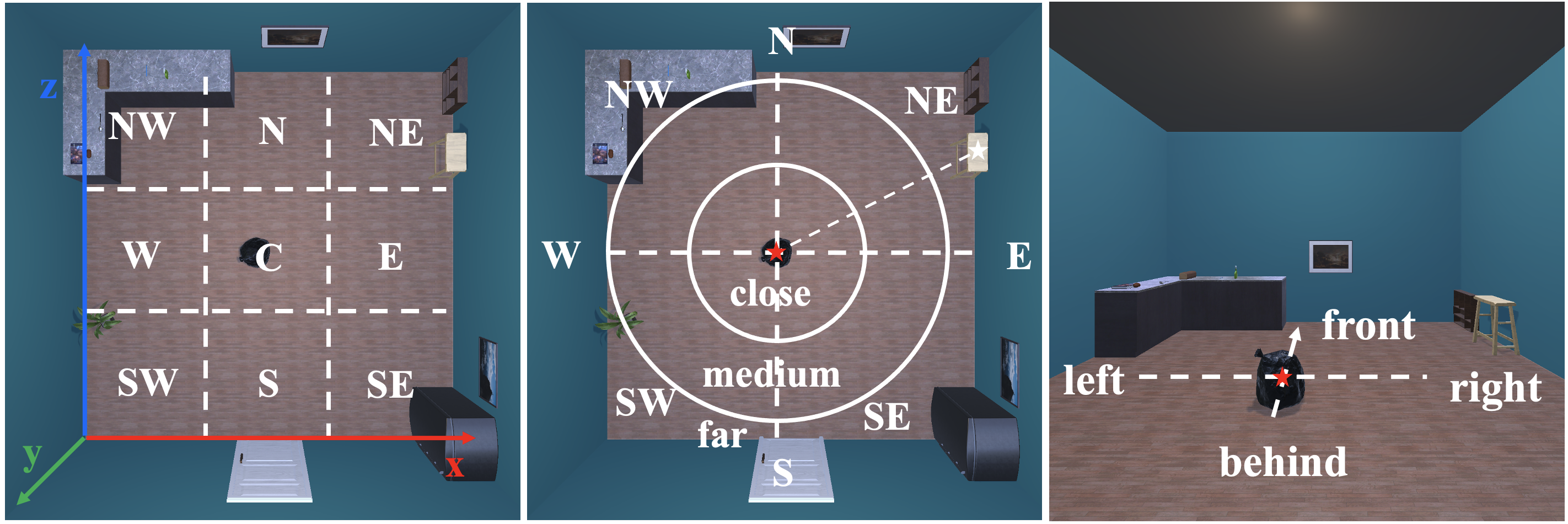}
    \caption{An overview of directional and distance spatial relationships: The left image displays the room's spatial divisions. The middle image displays both directional and distance-based relationships among objects from a top-down view.  The right image illustrates directional relations as seen from a north-facing perspective.}
    \label{fig:direction}
\end{figure}

\paragraph{Directional Relations.} 
We use a projection-based method to represent the nine different directional relations \fj{in} cardinal algebra \cite{ligozat1998reasoning}, as illustrated in the middle part of \fj{Figure} \ref{fig:direction}. 
We use two reference frames: \textbf{top-down view} and \textbf{north-facing view}, differing in the expression of binary-directional relations.
In the top-down view, these relations are depicted using cardinal directions (\textit{north}, \textit{south}, \textit{east}, \textit{west}) and their combinations.
In the facing view, \fj{the} cardinal directions are adapted to localized terms (\textit{front}, \textit{behind}, \textit{right}, \textit{left}) to provide a \fj{potentially} more intuitive understanding of spatial relations from the observer's viewpoint\footnote{\fj{It would not be intuitive in the aboriginal language \emph{Guugu Yimithirr}, which lacks words for `left' or `right', and spatial information is mainly conveyed using cardinal directions \cite{haviland1998guugu}.} }.

\paragraph{Distance Relations.}
The distance between objects is determined by calculating the Euclidean distance between the center points of their bounding boxes \( dis = \sqrt{(x_1-x_2)^2 + (z_1-z_2)^2} \). The qualitative distance relations are defined based on the ratio \( \frac{dis}{w} \) or \( \frac{dis}{\sqrt{2}w} \), where \( w \) is the length and width of the \tony{square} room,  \( \sqrt{2}w \) corresponds to the diagonal length of the room. We have incorporated two levels of distance relation settings in our benchmark:

\begin{itemize}[leftmargin=*]
    \item \textit{close}, \textit{far} (Threshold: $\frac{w}{2}$). A binary classification where \textit{close} is within half the room's width/length $w$, and \textit{far} is beyond it, providing a simple
    distance distinction.
    \item \textit{close}, \textit{medium}, \textit{far} (Thresholds: $\frac{\sqrt{2}w}{3}$, $\frac{2\sqrt{2}w}{3}$).  The \textit{medium} category is introduced for a more nuanced understanding, with \textit{close} up to $\frac{\sqrt{2}w}{3}$, \textit{medium} between $\frac{\sqrt{2}w}{3}$ and $\frac{2\sqrt{2}w}{3}$, \textit{far} beyond $\frac{2\sqrt{2}w}{3}$, as depicted in the middle part of Figure \ref{fig:direction}.
\end{itemize}



\subsection{CSP Example Generation}
\label{decide_p}

\subsubsection{Building  a Constraint Graph}


Our benchmark offers a variety of stories with varying levels of complexity, accomplished by adjusting two key parameters: \( n \) for object selection and \( m \) for constraint determination. Our methodology is implemented as follows:

\paragraph{Node Selection.} 
We focus on prominent, larger objects that occupy more space in a room. For example, in the context of `an apple on a desk', we would prioritize the desk over the apple. 
Of the \( N \) prominent objects in the scene, we randomly select \( n \)  \fj{to represent as} nodes in the graph.

\paragraph{Constraint Selection.}
In a constraint graph with \( n \) objects, there are \( C_n^2 \) potential pair connections. For example, a graph with 5 objects yields \( C_5^2 = 10 \) possible constraint pairs. 
For all possible pairs of objects, we first select one pair to form the question. Then, for the remaining \( C_n^2-1\)  pairs, the parameter \( m \) is used to establish graph. 





\subsubsection{Answer - Consistency Checking}

\fj{We include two types of questions:
\textbf{Find Relation (FR)}: identify the directional spatial relationship between two specified objects. \textbf{Yes/No (YN)}: ascertain the validity of a statement concerning the spatial relationship between objects.}

\begin{figure*}[htb]
    \centering
    \includegraphics[width=1\textwidth]{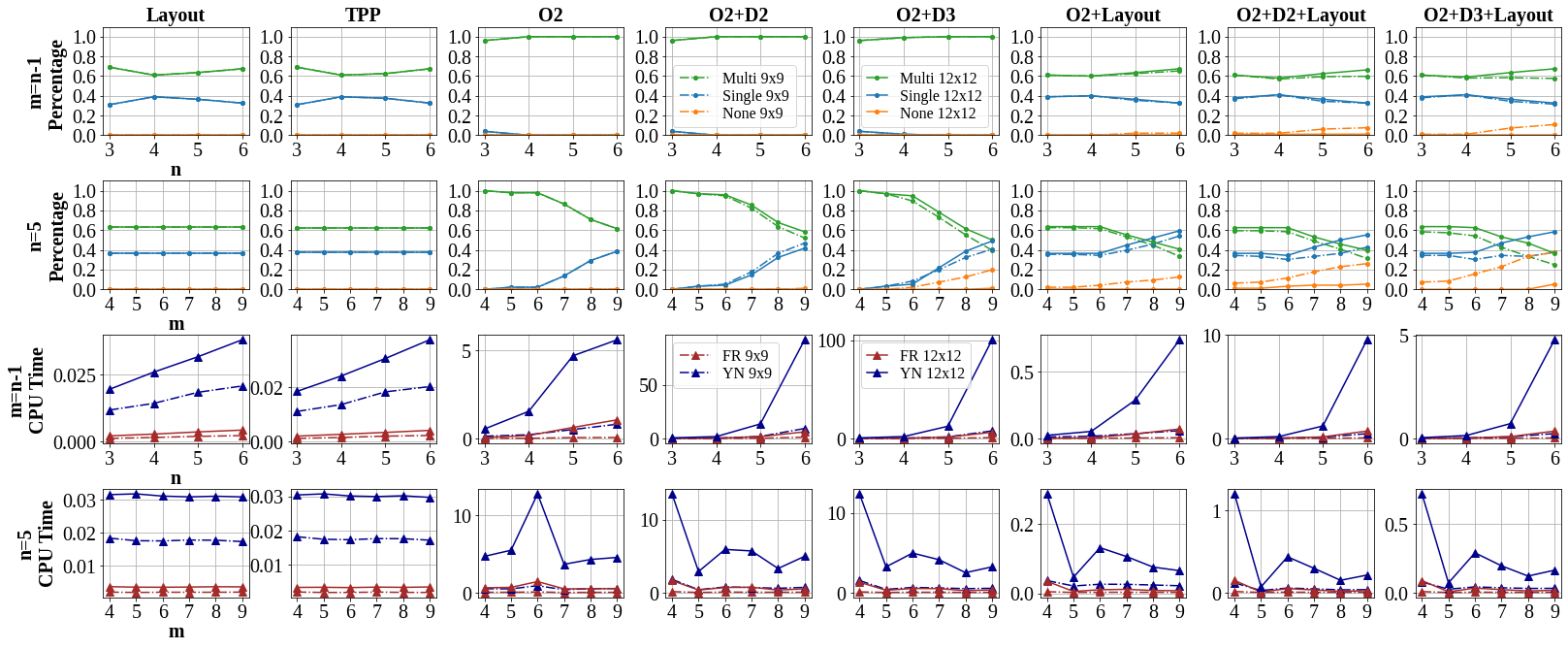}
    \caption{
    The percentage of \textit{single}, \textit{multiple}, and \textit{no} solution occurrences (Rows 1, 2) and the average CPU time \fj{(seconds)} for solution searches (Rows 3, 4) in RoomSpace-100 with different $d$. For Rows 1 and 3, \( n \) varies while \( m = n-1 \); for Rows 3 and 4, \( m \) varies with \( n \) constant at 5. Spatial relation settings include \textbf{Layout}: The basic setting with directional object layout relations. \textbf{TPP}: Enhanced object layout with topological relations TPP and NTPP. \textbf{O2}: Pure inter-object directional relations. \textbf{O2+D2}: O2 expanded with two distance relations; \textbf{O2+D3}: O2 expanded with three distance relations; \textbf{O2+D2+Layout} and \textbf{O2+D3+Layout}: Combining inter-objects relations with object layout relations.
    }
    \label{fig:dataset_plot1}
\end{figure*}

Generating ground-truth answers for spatial relations between objects $o_1$ and $o_2$ from the simulation system can be automated through comparing their coordinates, represented as \((x_1, y_1)\) and \((x_2, y_2)\). 
However, key considerations arise: Given the stories formed with limited qualitative relations, can we definitively deduce the answer? Is there a possibility of multiple valid solutions? For example, in the scenario `A is to the left of B, and C is to the left of A,' the position of A relative to C is ambiguous based on the information provided. A could be to the right, left, or overlapping with C. 
The stories in our benchmark offer a partial view of spatial layouts. 
Given the limited qualitative descriptions, a singular, definitive answer may not always be attainable.

Recognizing the potential for multiple valid solutions within the constraints detailed in the story, we have developed a consistency-checking tool using the \textbf{python-constraint} package\footnote{https://github.com/python-constraint/python-constraint}, which employs \fj{a} backtracking algorithm to determine whether a plausible configuration of object relationships can exist to meet all specified constraints. Additional information about this reasoner is available in the Appendix.

In Figure \ref{fig:dataset_plot1}, we analyze the occurrence of \textit{single}, \textit{multiple}, and \textit{no} solution possibilities under various constraint settings.
With a smaller domain size of $9 \times 9$, the \textit{Layout} and \textit{O2} relation settings consistently yield solutions; however, the likelihood of \textit{no} solution is significantly higher compared to the larger domain size of $12 \times 12$ when incorporating distance constraints. 
Additionally, the search cost (CPU time) required to find solutions with the larger domain size is considerably higher than with the smaller one.
We examine the search costs associated with finding solutions for FR and YN questions. FR questions generally involve multiple answers and require evaluating all nine \fj{direction relations} to identify all potential solutions that meet the constraints. In contrast, YN questions involve checking only one relational candidate, resulting in lower search costs.




\subsection{Generate Textual Descriptions}
\label{decide_t}

During this phase, we transform the spatial logical expressions \(C_l\) and \(C_o\) into natural language sentences \(S_l\) and \(S_o\), a process known as logic-to-text generation.

We develop specific logic-to-string templates using context-free grammar (CFG). 
When forming stories, the logical components such as \textit{\(\langle x_i\rangle \),\(\langle x_j \rangle \), \(\langle r_i^{Dir} \rangle \),\(\langle r_i^{TPP} \rangle \), \(\langle r_{ij}^{Dir} \rangle \), \(\langle r_{ij}^{Dis} \rangle \)} are replaced with corresponding textual expressions, enabling the creation of varied descriptions of spatial relationships. 
Our CFG has two parts, as shown in Table \ref{tab:CFG}.

\begin{table}[tb]
    \setlength{\tabcolsep}{1.0mm} 
    \centering
    \begin{tabular}{|p{8.3cm}|}
    \hline
      \textit{$S_{l} \rightarrow$ This room contains a collection of furniture, including $\langle S_{l}^{0} \rangle$, $\langle S_{l}^{1} \rangle$, \dots , $\langle S_{l}^{n} \rangle$.}\\
       \textit{$S_{o}^{T} \rightarrow$ $\langle S_{o}^{T01} \rangle$. $\langle S_{o}^{T12} \rangle$. \dots . $\langle S_{ot}^{Tij} \rangle$. }\\
       \textit{$S_{o}^{N} \rightarrow$ Imagine yourself at the southern wall's door, looking inwards. From this perspective, $\langle S_{o}^{N01} \rangle$. \dots .$\langle S_{o}^{Nij} \rangle$. }\\
       \\
       \textit{$S_{l}^{i} \rightarrow \langle x_i \rangle $ placed in the $\langle r_{i}^{Dir} \rangle$, $\langle r_i^{TPP} \rangle $ the wall}\\
       \textit{$S_{o}^{Tij} \rightarrow \langle x_i \rangle $ is placed to the $\langle r_{ij}^{Dir} \rangle $ of $\langle x_j \rangle $, $\langle r_{ij}^{Dis} \rangle$}\\
        \textit{$ S_{o}^{Nij} \rightarrow \langle x_i \rangle $ is $\langle r_{ij}^{Dir\_N} \rangle $ $\langle x_j \rangle $, $\langle r_{ij}^{Dis} \rangle$.}\\
        \hline
    \end{tabular}
    \caption{Our designed grammar. \(S^{N}\) represents sentences describing north-facing view relations, and \(S^{T}\) for top-down views. }
    \label{tab:CFG}
\end{table}

\section{Evaluation}
\subsection{Model Settings and Prompting}
We access GPT-3 (Davinci) \cite{brown2020language}, \fj{GPT-3.5 (Turbo)}, and GPT-4 \cite{OpenAI2023GPT4TR} via the Azure OpenAI Service, using the API version ``\textit{2023-09-15-preview}" for all three models. To yield more deterministic results, we set the temperature to 0 in all experiments.
The remaining parameters were left at the standard configurations for these models.

\label{section:prompting}
We conduct experiments with two sets of prompts \cite{bommasani2021opportunities}: one set directly presents stories and questions to LLMs, while the other incorporates task descriptions and details about relationship definitions, as detailed in the Appendix, to guide LLMs' responses.



\fj{Experiment results (in Appendix)} illustrates a slight improvement in the performance of \textit{gpt-35-turbo} with the \textit{Layout, O2+D2, and O2+D2+Layout} settings. However, incorporating task description prompts results in a decrease in accuracy within the \textit{TPP} settings. Therefore, although the added prompts about task description provide valuable insights into the spatial reasoning problem, the minimal variation in performance suggests that for subsequent experiments, we  maintain a straightforward story and question format prompt.

\subsection{Results}

\begin{figure}[tb]
    \centering
    \includegraphics[width=0.48\textwidth]{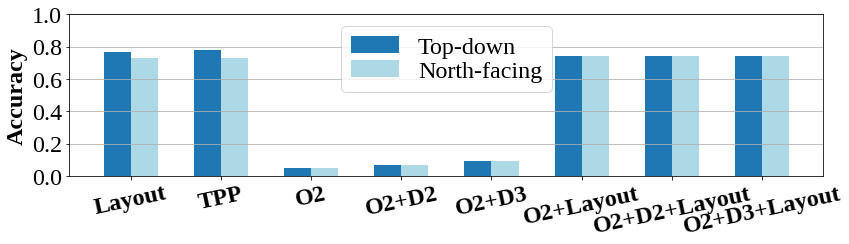}
    \caption{Performance of \textit{gpt-35-turbo} on the \textit{RoomSpace-100} test sets with \( n=5 \) and \( m=4 \) using top-down view and north-facing view on YN questions. 
    }
    \label{fig:accuracy_2}
\end{figure}

\begin{figure}[tb]
    \centering
    \includegraphics[width=0.48\textwidth]{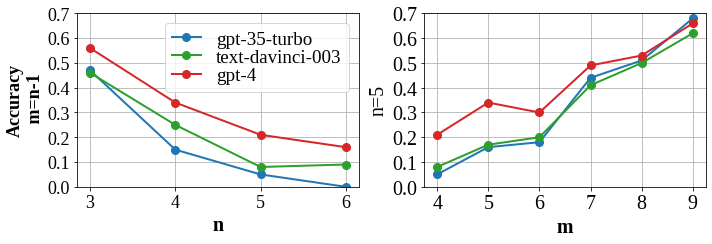}
    \caption{Performance of GPT models with top-down view \textit{O2} setting across variations in parameters \( n \) and \( m \) on \textit{RoomSpace-100}. }
    \label{fig:p1p2}
\end{figure}

Figure \ref{fig:accuracy_2} and Figure \ref{fig:p1p2} present the comparative results across models, relation settings, parameters \fj{\( n \) and \( m \)}, highlighting several key observations:

\paragraph{Model Comparison.}
GPT-4 consistently surpasses both Turbo and Davinci in nearly all categories and from various viewpoints. Turbo shows comparatively lower accuracy than the other two models, with its accuracy falling to zero under the condition where \( n=6 \) and \( m=5 \). 

\paragraph{Viewing Perspective Influence.}
The north-facing view descriptions do not significantly impact the results when the narrative already includes descriptions from that view, as in the \textit{O2 }setting and its combinations with distance or layout, where accuracy remains comparable to the top-down view.  However, under the \textit{Layout} setting, which includes directional descriptions from the top-down view, introducing north-facing view descriptions in the questions complicates comprehension for LLMs, leading to a decline in accuracy.

\paragraph{Impact of Spatial Reasoning Settings.}

\textit{Layout} vs. \textit{O2}: 
In the Layout setting, the introduction of TPP does not markedly affect accuracy. Even with \( n=5 \), GPT consistently performs well, efficiently extracting and analyzing information. However, when dealing with only the relationships between objects in multi-object scenes, the task becomes challenging for GPT, highlighting the model's limitations in multi-hop spatial reasoning.
   
Distance Settings (\textit{D2, D3}): Interestingly, Turbo’s performance slightly improves with the introduction of distance constraints. This may suggest GPT-4's better handling of more complex spatial relations.
   
Combination of \textit{Layout}, \textit{O2} and Distance: The combined settings typically yield performance that is on par with the best-performing individual setting, in this instance, aligning with the results observed in the layout setting.


\paragraph{Variation with Parameters (\( n \) and \( m \)).}
There is a decline in accuracy as \( n \) increases from 3 to 7, suggesting that larger \( n \) values create more complex and challenging scenarios  \fj{(see Figure \ref{fig:p1p2}, left)}. This trend aligns with the observations in Figure \ref{fig:dataset_plot1} -
the time taken by the CPU to find solutions increases with higher \( n \) values. 
In terms of \( m \), an increase in this parameter generally leads to improved accuracy \fj{(see Figure \ref{fig:p1p2}, right)}. It appears that larger \( m \) values, with more densely interlinked spatial relationships, though adding text length, tend to enhance LMs' performance.

\section*{Conclusion}

Our study identifies gaps in current QSR datasets and presents a new benchmark to better evaluate LMs' capabilities in spatial reasoning. We enhance QSR dataset creation with a benchmark that addresses multiple complexities, including topological, directional, and distance relationships. Our benchmark uniquely incorporates different viewing perspectives in spatial reasoning, moving towards more accurate LM evaluations. Our results underscore the necessity for enhancements in current state-of-the-art LLMs, opening new avenues for enhancing spatial reasoning in AI models.

Future directions include incorporating object size and shape, as our current focus is on object centers for spatial relationships. Additionally, exploring more topological relations beyond TPP and NTPP can deepen the benchmark's scope. We also aim to include more complex perspectives, such as an agent's viewpoint within a room, introducing natural front-facing scenarios for more challenging reasoning tasks.

This paper provides a preliminary evaluation of OpenAI's GPT series models \fj{on our new dataset \textit{RoomSpace-100}}. Expanding this research to assess and compare the spatial reasoning abilities of other LLMs would be beneficial.  Additionally, although our benchmark covers both FR and YN questions, our evaluation is limited to the YN questions.  FR questions, which typically require multiple-choice answers, represent a more significant challenge. Future research could delve into these more intricate scenarios. Moreover, while our evaluations utilize \textit{RoomSpace-100}, exploring larger sets, such as the 1K and 10K versions, could provide more comprehensive insights.

\section*{Acknowledgments}
We thanks the anonymous referees for their helpful comments. This work has been partially supported by: (1) Microsoft Research - Accelerating Foundation Models Research program, with the provision of Azure resources to access GPT; (2) the Turing’s Defence and Security programme through a partnership with the UK government in accordance with the framework agreement between GCHQ and The Alan Turing Institute; (3)  Economic and Social Research Council (ESRC) under grant ES/W003473/1.

\section*{Data Access Statement}
Data associated with this paper are available from the University of Leeds data repository \url{https://doi.org/10.5518/1518}. Code and appendix are available at \url{https://github.com/Fangjun-Li/RoomSpace}.

\section*{Author Contributions}
AC conceived the original idea for the benchmark which was then refined in discussions with FJ and DH. FJ implemented the benchmark and designed all details, performed the evaluations, and wrote the original draft of the paper. All authors contributed to the subsequent drafts.

\bibliographystyle{named}
\bibliography{ijcai24}

\clearpage
\appendix

\section{Existing Benchmarks}
\begin{figure}[ht]
    \centering
    \includegraphics[width=0.48\textwidth]{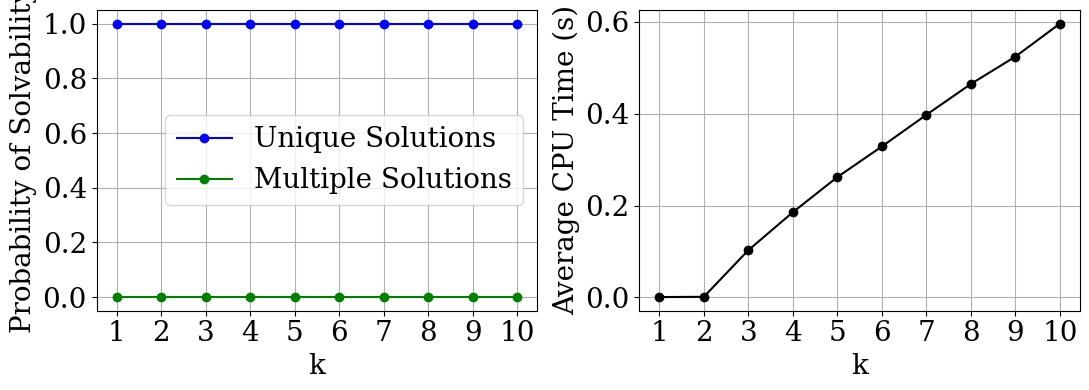}
    \caption{The probability of solvability
    (determining if there is a unique solution or multiple solutions) and the average CPU time to find a solution for each level of $k$.
    }
    \label{fig:complexity01}
\end{figure}

\begin{table}[ht]
    \centering
    \small
    \setlength{\tabcolsep}{0.5mm}
    \begin{tabular}{p{1.5cm}|p{2.3cm}| p{1.5cm}|p{2.8cm}}
    \toprule
     & \textbf{bAbI} & \textbf{StepGame} & \textbf{SpartQA} \\
    \midrule
    Represent &  \multicolumn{2}{c|}{Grid-based (point)}  & shape-based \\
    \hline
    object type & color+shape(T17), location(T19)& A-Z & shape + size + color \\ 
    \hline
    relation & relative(T17), \ \ \ cardinal(T19) & relative, cardinal & cardinal, distance, RCC,   \\
    \hline
    constraint & \multicolumn{2}{c|}{unit distance and angle} & relation defination\\ 
    \hline
    description & templates & templates & CFG \\ 
    \hline
    Images & \multicolumn{2}{c|}{ No} & Yes, but not matching the story's objects.\\ 
    \bottomrule
    \end{tabular}
    \caption{Comparison of existing QSR in text datasets}
    \label{tab:my_label}
\end{table}

\section{Logical Reasoner for Consistency Checking}


Our CSP is represented in terms of variables (objects), finite room domains, and constraints (spatial relation sets).

\begin{figure}[ht]
    \centering
    \includegraphics[width=0.48\textwidth]{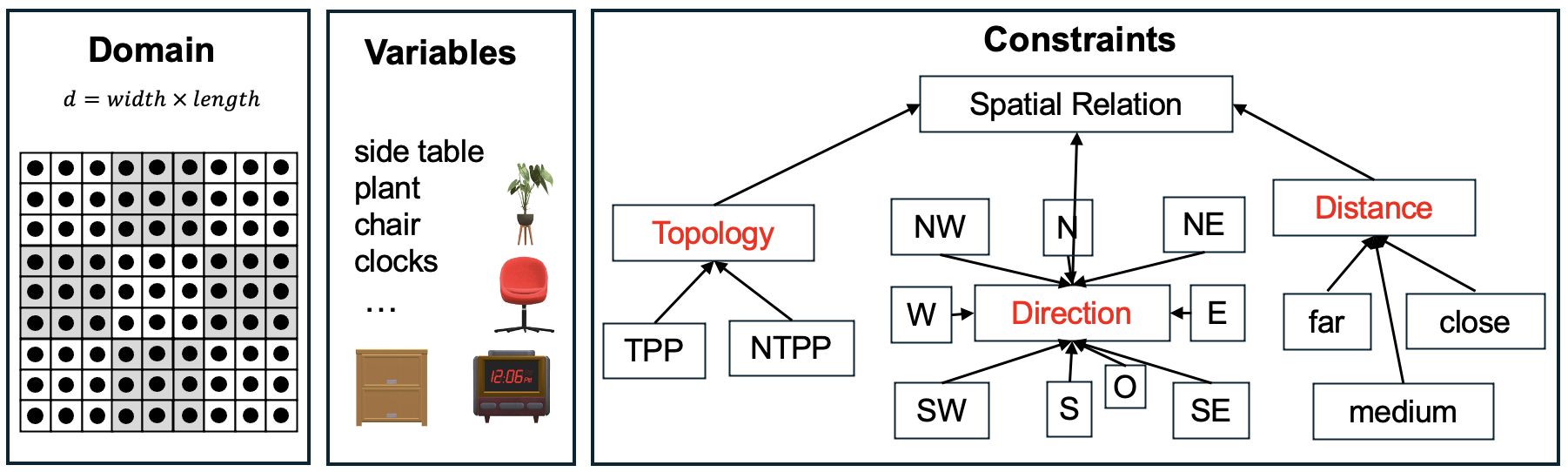}
    \caption{Visualization of our CSP frameworks. The domain size shown here (\(d = 9 \times 9\)) is for illustrative purposes only.  
    }
    \label{fig:A1}
\end{figure}


\begin{figure}[htb]
    \centering
    \includegraphics[width=0.5\textwidth]{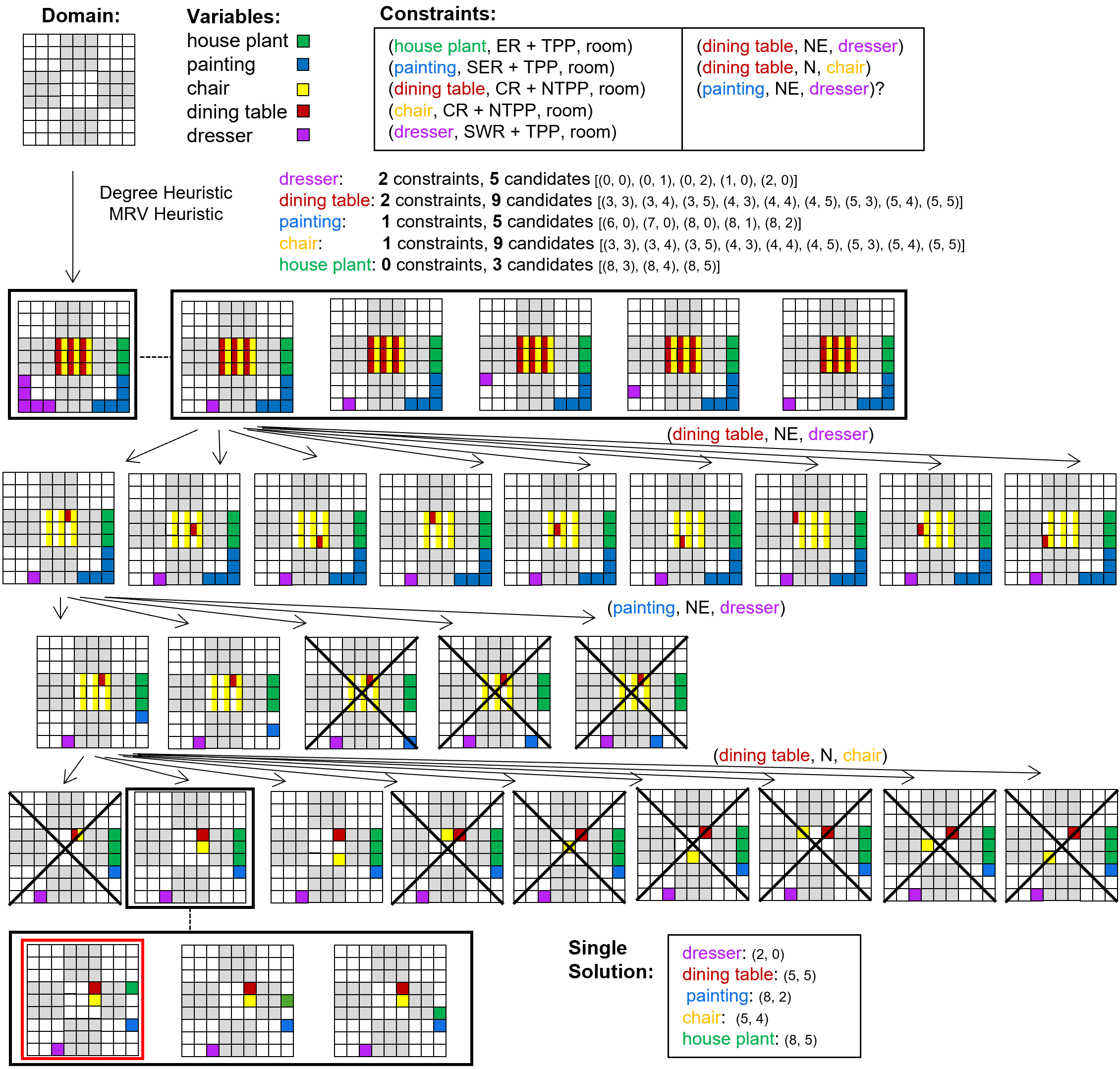}
    \caption{Example backtracking solving process to get a single solution with finite domain $ d= 9 \times 9$, $n=5$ variables, five constraints about object layouts and three constraints about relations between objects.}
    \label{fig:A3}
\end{figure}
We use the backtracking algorithm for solving our CSP,
where the goal is to find an assignment of values to variables that satisfies all given constraints. Here is a 
\tony{summary}
of how the algorithm works:

\begin{enumerate} [leftmargin=*]
\setlength{\itemsep}{0pt}
\setlength{\parsep}{0pt}
\setlength{\parskip}{0pt}
\item \textbf{Variable Selection}: Employs a combination of the Degree Heuristic (prioritize variables with the most constraints on remaining variables) and the Minimum Remaining Values (MRV) heuristic (prioritize variables with the fewest legal values left) to choose the next variable to assign. This helps in reducing the search space and picking the most constrained variable first.
\item \textbf{Value Selection and Forward Checking}: For the selected variable, iterate through its possible values (domain). Forward checking tentatively assigns a value and checks if this leads to any immediate dead ends in the remaining variables (i.e., if any variable is left with no possible values), which helps in pruning the search space early. 
\item \textbf{Constraint Checking}: For each value attempted, check all relevant constraints. If a value does not satisfy a constraint, it is discarded, and the algorithm tries the next value. If all values are tried and none fits, backtrack to the previous variable and try a new value for it, undoing any changes made since that variable was assigned.
\item \textbf{Solution Yielding and Backtracking}: Once all variables are assigned in a way that all constraints are satisfied, a solution is yielded. If the solution space is exhausted for the current path, backtrack to explore other paths.
\end{enumerate}


 




\subsection{Complexity}


The time complexity of backtracking algorithms for CSPs can vary significantly based on the problem's constraints, the size of the domains and variables, and the heuristics used:

\textbf{Worst-Case Time Complexity}: In the worst case, the algorithm might explore every possible assignment of values to variables, leading to a time complexity of \(O(d^n)\), where \(d= 3\times3\) is the size of the largest domain, and \(n\) is the number of variables.

\subsection{Influence of Constraints}
The number and type of constraints significantly affect the complexity of solving a CSP. 
\tony{Here is} how constraints can impact the complexity:

\textbf{Tightness}: 
This refers to how restrictive a constraint is. A tighter constraint eliminates more values from the domains of the variables it involves.  
To illustrate, consider the constraint (house plant, ER, room) applied to the variable `house plant' within a domain of $9=3\times3$ possibilities. Introducing constraint (house plant, ER+TPP, room) further reduces the domain to only 5 possible pairs: [(0, 0), (0, 1), (0, 2), (1, 0), (2, 0)]. 
This is the reason, as depicted in Figure \ref{fig:dataset_plot1}, that there is a marked decrease in CPU times when we transition from a `\textit{Layout}' configuration to a `\textit{TPP}' setting.


Tighter constraints can reduce the search space because fewer value combinations are valid, but they also increase the chance of running into a dead end (where no solution is possible), which can force more backtracking.
Figure \ref{fig:dataset_plot1} demonstrates that incorporating binary distance relations `\textit{O2+D2}' leads to a rise in the average CPU time when contrasted with the `\textit{O2}' configuration.

We provide an analysis of the constraint tightness \( p \) for all spatial relations. In our CSP solver, we explore two settings for domain size: \( d = 9 \times 9 \) and \( d = 12 \times 12 \), corresponding to the room's square configuration.

\textbf{Layout Constraints}:
These constraints are not represented in constraint graphs; rather, they are utilized directly to refine the domain of the variable they constrain. Though in the form \((x_i, R_i, Room) \), they function as unary constraints.

\begin{itemize}[leftmargin=*]
\item \textbf{InR} (In Room): \( p = 0 \), all possible values from the domain of the one variable are allowed and the constraint is always satisfied.

\item \textbf{NR, SR, WR, ER, NER, NWR, SER, SWR, CR}: \( p = \frac{8}{9}\), Each relation pertains to a specific section of the room, dividing the room into nine parts.

\item \textbf{TPP, NTPP}: TPP corresponds to the border of the grid space, with \( p^{TPP}  = \frac{(\sqrt{d}-2)^2}{d} \),
NTPP corresponds to the inner side of the room, with \( p^{NTPP} = \frac{(\sqrt{d}-1)\times 4}{d}  \)

\end{itemize}

\textbf{Inter-Objects Constraints}:
We use binary constraints involving two variables to represent the relationships between objects, which can be illustrated in constraint graphs.

\begin{itemize}[leftmargin=*]
\item \textbf{N, S, W, E, NE, NW, SE, SWR, O}: Directions between Objects.
For  N, S, W, E, \( p  = 1 -\frac{d(\sqrt{d}-1)}{2 d^2}\), for  NE, NW, SE, SWR, \( p = 1- ({\frac{d -\sqrt{d}}{2d} })^2 \). For O, \( p = 1-\frac{1}{d}\)

\item \textbf{CL2, FR2}: 
Objects are considered close (CL2) if they are within half of the maximum distances for one dimension (width or length). We approximate this using a circle with radius \( r_1=\frac{\sqrt{d}-1}{2} \), so  \({Area}^{CL2} = \pi (\frac{\sqrt{d}-1}{2})^2 \),
\({Area}^{FR2} = d-\pi (\frac{\sqrt{d}-1}{2})^2 \)
The \( p\) calculation for distance in terms of the grid dimension \( d\) is complex.
The number of cells within this area (\(p^{FR2}\)) for the central object can be approximated by:
\( p^{FR2} \approx \frac{{Area}^{CL2}}{d} \), \( p^{CL2}  \approx 1-  \frac{{Area}^{CL2}}{d} \).
For each central object, the actual count of possible variable values is limited by the number of cells that fit into this area.

\item \textbf{CL3, MD3, FR3}:
more restrictive than the previous two-category distances. Objects are considered close (CL3) if they are within one-third of the maximum distances within the grid, and moderate distance (MD3) if within two-thirds of the maximum distances within the grid.  We approximate this using a circle with radius \( r_1 = \frac{\sqrt{2}(\sqrt{d}-1)}{3} \), \( r_2 = \frac{2\sqrt{2}(\sqrt{d}-1)}{3} \).
\({Area}^{CL3} = \pi r_1^2 \),  \({Area}^{MD3} = \pi (r_2^2-r_1^2)  \)
\({Area}^{FR3} = d-\pi r_2^2  \).

\( p^{FR3} \approx 1- \frac{{Area}^{FR3}}{d} \), 
\( p^{MD3}  \approx 1-  \frac{{Area}^{MD3}}{d} \)
\( p^{CL3}  \approx 1-  \frac{{Area}^{CL3}}{d} \)
.
\end{itemize}

\textbf{Density}: This is about the number of constraints in the problem relative to the number of variables, aligning with \(m\) in Figure \ref{fig:dataset_plot1}. A higher density means more pairs of variables are constrained relative to each other. More constraints generally increase the complexity because more conditions must be satisfied, potentially increasing the amount of backtracking required.

In Figure \ref{fig:dataset_plot1}, There is a rising curve of CPU times as \(n\) increases across all six relational configurations.
Theoretically, the worst-case time complexity is expected to escalate exponentially. However, due to forward checking and the use of MRV and Degree heuristics, the effective branching factor is often significantly reduced, leading to much better performance in practice.



\section{Prompt Design}

The prompts mentioned in Section \ref{section:prompting} were designed to clarify spatial relations, reducing ambiguity in spatial descriptions.
For different sets of spatial reasoning problems, corresponding guidelines will be combined with the task description part to form the prompt.

\begin{figure}[htb]
    \centering
    \includegraphics[width=0.48\textwidth]{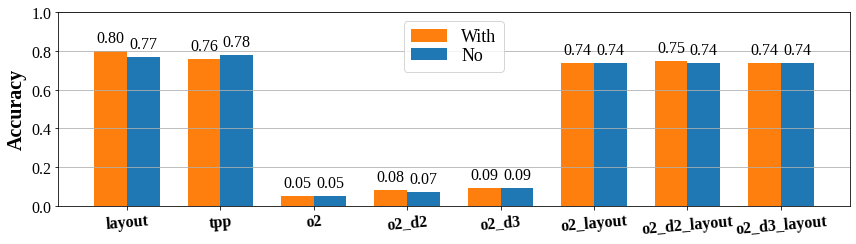}
    \caption{Performance of \textit{gpt-35-turbo} on the \textit{RoomSpace-100} test sets with \( n=5 \) and \( m=4 \) using top-down view YN questions. 
    The `\textit{No}' bar shows results obtained without introductory prompts; the `\textit{With}' bar presents results with introductory prompts included.
    }
    \label{fig:accuracy_1}
\end{figure}

\begin{itemize} [leftmargin=*]
    \item \textbf{Task}: \textit{Analyze the spatial relationships between specified objects in a room, treating each object as a point within a 12$\times$12 grid.}
    \item \textbf{Distance-2}: \textit{Distances between objects in the room are determined using the room's width. A `short distance' is defined as any distance up to half of the room's width. A `far distance' refers to any distance that exceeds half of the room's width.}
    \item \textbf{Distance-3}: \textit{Distances between objects in the room are determined based on the room’s diagonal length. A `short distance' refers to a distance that is up to one-third of the diagonal. A `moderate distance' spans from one-third to two-thirds of the diagonal. A `far distance' is any distance that exceeds two-thirds of the diagonal.}
\end{itemize}

\end{document}